\documentclass[10pt,twocolumn,letterpaper]{article}

\usepackage{cvpr}
\usepackage{times}
\usepackage{epsfig}
\usepackage{graphicx}
\usepackage{amsmath}
\usepackage{amssymb}

\usepackage[pagebackref=true,breaklinks=true,letterpaper=true,colorlinks,bookmarks=false]{hyperref}

\cvprfinalcopy 

\def\cvprPaperID{729} 

\ifcvprfinal\fi
\begin{document}

\title{Person Search with Natural Language Description}

\author{Shuang Li$^{1}$\ \ \ \ Tong Xiao$^{1}$\ \ \ \ Hongsheng Li$^{1*}$\ \ \ \ Bolei Zhou$^{2}$\ \ \ \ Dayu Yue$^{3}$\ \ \ \ Xiaogang Wang$^{1}$\thanks{Corresponding authors}\\
\small $^{1}$The Chinese University of Hong Kong \
\small $^{2}$Massachuate Institute of Technology\
\small $^{3}$SenseTime Group Limited\\
{\tt\small \{sli,xiaotong,hsli,xgwang\}@ee.cuhk.edu.hk,\ bolei@mit.edu,\ yuedayu@sensetime.com}
}

\maketitle
\thispagestyle{empty}

\begin{abstract}
Searching persons in large-scale image databases with the query of natural language description has important applications in video surveillance. Existing methods mainly focused on searching persons with image-based or attribute-based queries, which have major limitations for a practical usage. In this paper, we study the problem of person search with natural language description. Given the textual description of a person, the algorithm of the person search is required to rank all the samples in the person database then retrieve the most relevant sample corresponding to the queried description. Since there is no person dataset or benchmark with textual description available, we collect a large-scale person description dataset with detailed natural language annotations and person samples from various sources, termed as CUHK Person Description Dataset (CUHK-PEDES). A wide range of possible models and baselines have been evaluated and compared on the person search benchmark. An Recurrent Neural Network with Gated Neural Attention mechanism (GNA-RNN) is proposed to establish the state-of-the art performance on person search.

\end{abstract}

\begin{figure}
\begin{center}
\includegraphics[width=0.95\linewidth]{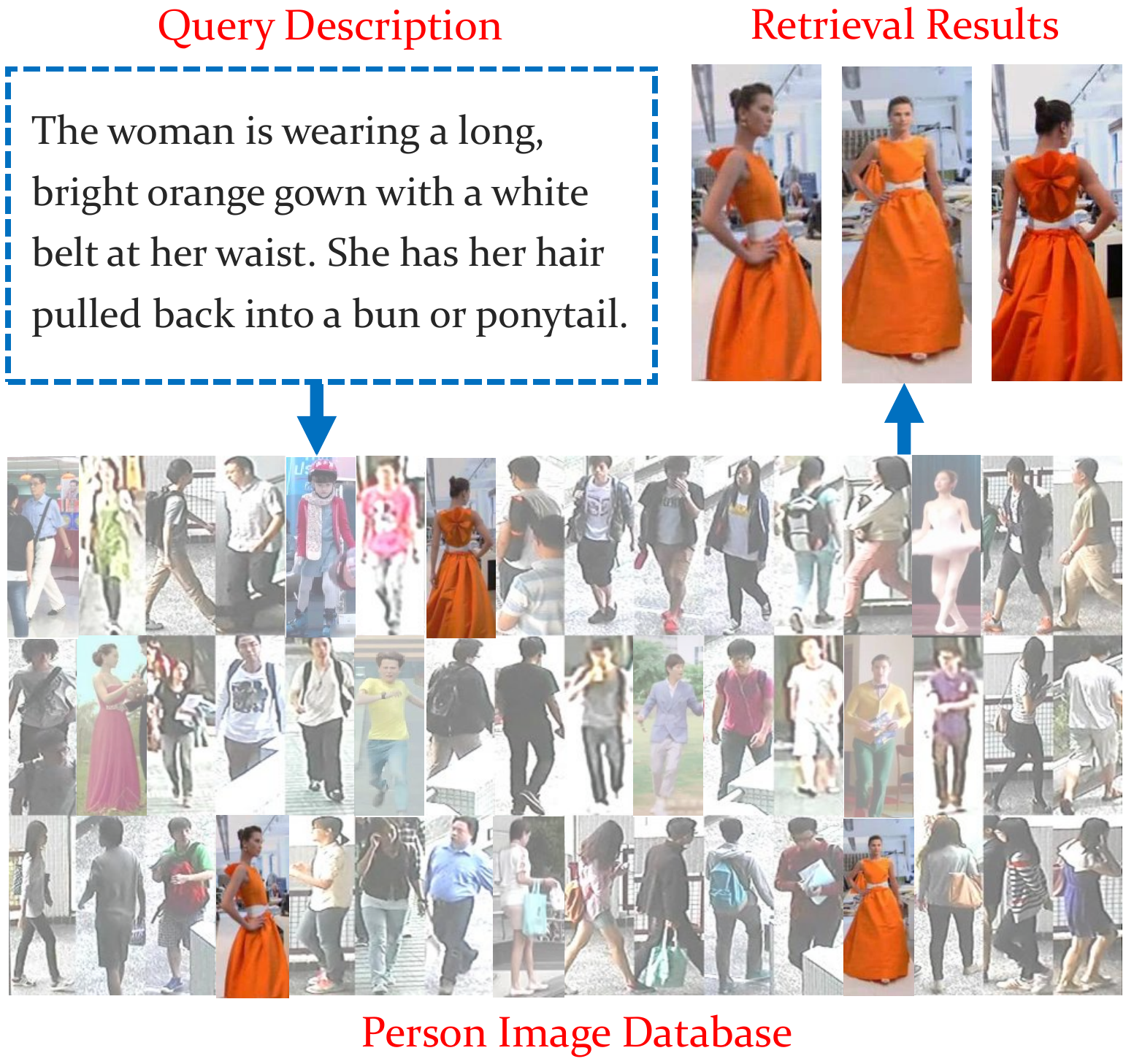} \ \\
\end{center}
\vspace{-8pt}
\caption{Given the natural language description of a person, our person search system searches through a large-scale person database then retrieve the most relevant person samples. }
\vspace{-7pt}
\label{fig:demo}
\end{figure}

\begin{figure*}
\begin{center}
\includegraphics[width=0.9\linewidth]{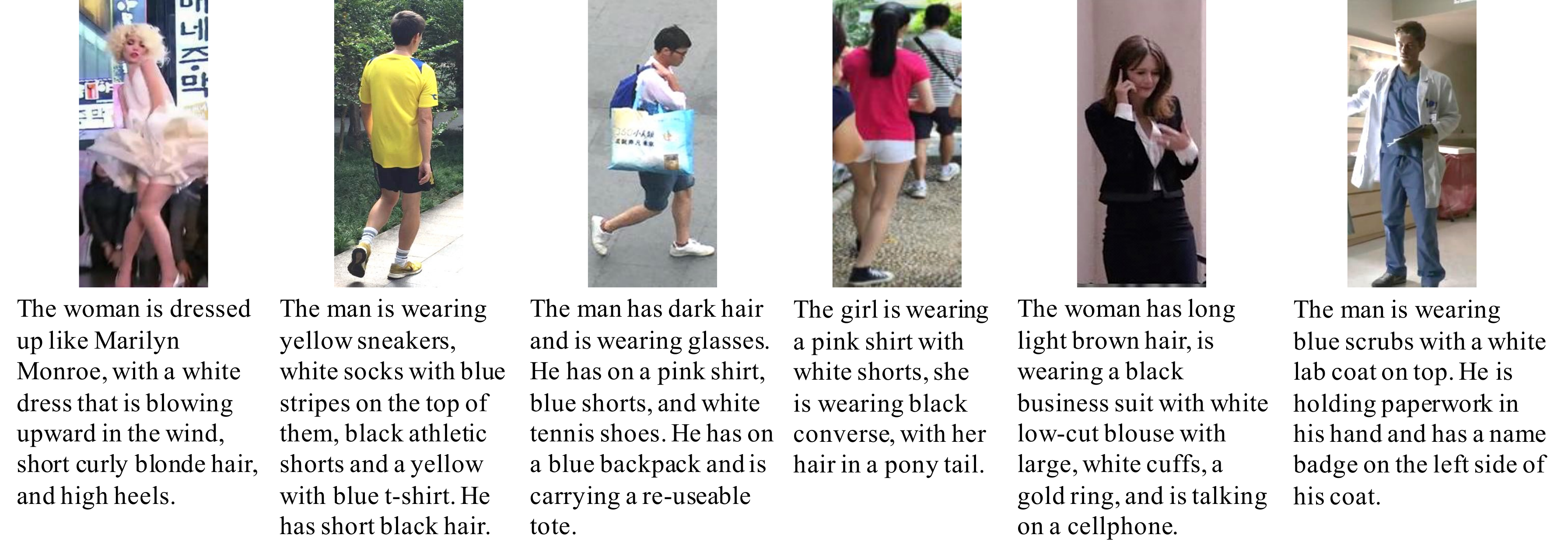} \ \\
\end{center}
\vspace{-8pt}
\caption{Example sentence descriptions from our dataset that describe persons' appearances in detail.}
\label{fig:dataset}
\vspace{-5pt}
\end{figure*}

\section{Introduction}
Searching person in a database with free-form natural language description is a challenging problem in computer vision. It has wide applications in video surveillance and activity analysis. Nowadays urban areas are usually equipped with thousands of surveillance cameras which generate gigabytes of video data every second. To search possible criminal suspects from such large-scale videos manually might take tens of days or even months to complete. Thus automatic person search is in urgent need. Based on modalities of the queries, existing person search methods can be mainly categorized into the ones with image-based queries and attribute-based queries. However, both modalities have major limitations and might not be suitable for practical usages. Facing such limitations, we propose to study the problem of searching persons with natural language descriptions. Figure \ref{fig:demo} illustrates one example of the person search.


Person search with image-based queries is known as person re-identification in computer vision \cite{zheng2011person,liao2015person,xiao2016end}. Given a query image, the algorithms obtain affinities between the query and those in the image database. The most similar persons can be retrieved from the database according to the affinity values. However, such a problem setting has major limitations in practice, as it requires at least one photo of the queried person being given. In many criminal cases, there might be only verbal description of the suspects' appearance available.

Person search could also be done through attribute-based queries. A set of pre-defined semantic attributes are used to describe persons' appearances. Classifiers are then trained on each of the attributes. Given a query, similar persons in the database can be retrieved as the ones with similar attributes \cite{vaquero2009attribute,su2016deep}. However, the attributes have many practical limitations as well. On the one hand, attributes have limited capability of describing persons' appearance. For instance, the PETA dataset \cite{deng2014pedestrian} defined 61 binary and 4 multi-class person attributes, while there are hundreds of words for describing a person's appearance. On the other hand, even with the exhausted set of attributes, labeling them for a large-scale person image dataset is expensive. 

Facing the limitations of both modalities, we propose to use natural language description to search person. It does not require a person photo to be given as in those image-based query methods. Natural language also can precisely describe the details of person appearance, and does not require labelers to go through the whole list of attributes. 

Since there is no existing dataset focusing on describing person appearances with natural language, we first build a large-scale language dataset, with 40,206 images of 13,003 persons from existing person re-identification datasets. Each person image is described with two sentences by two independent workers on Amazon Mechanical Turk (AMT). On the visual side, the person images pooled from various re-identification datasets are under different scenes, view points and camera specifications, which increases the image content diversity. On the language side, the dataset has 80,412 sentence descriptions, containing abundant vocabularies, phrases, and sentence patterns and structures. The labelers have no limitations on the languages for describing the persons. We perform a series of user studies on the dataset to show the rich expression of the language description. Examples from the dataset are shown in Figure \ref{fig:dataset}.

We propose a novel Recurrent Neural Network with Gated Neural Attention (GNA-RNN) for person search. The GNA-RNN takes a description sentence and a person image as input and outputs the affinity between them. The sentence is input into a word-LSTM and processed word by word. At each word, the LSTM generates unit-level attentions for individual visual units, each of which determines whether certain person semantic attributes or visual patterns exist in the input image. The visual-unit attention mechanism weights the contributions of different units for different words. In addition, we also learn word-level gates that estimate the importance of different words for adaptive word-level weighting.  The final affinity is obtained by averaging over all units' responses at all words. Both the unit-level attention and word-level sigmoid gates contribute to the good performance of our proposed GNA-RNN.

The contribution of this paper is three-fold. 1) We propose to study the problem of searching persons with natural language. This problem setting is more practical for real-world scenarios. To support this research direction, a large-scale person description dataset with rich language annotations is collected and the user study on the natural language description of person is given. 2) We investigate a wide range of plausible solutions based on different vision and language frameworks, including image captioning~\cite{karpathy2015deep, vinyals2015show}, visual QA~\cite{zhou2015simple,ren2015exploring}, and visual-semantic embedding~\cite{reed2016learning}, and establish baselines on the person search benchmark.  3) We further propose a novel Recurrent Neural Network with Gated Neural Attention (GNA-RNN) for person search, with the state-of-the-art performance on the person search benchmark. 


\subsection{Related work}

As there are no existing datasets and methods designed for the person search with natural language, we briefly survey the language datasets for various vision tasks, along with the deep language models for vision that can be used as possible solutions for this problem.

\textbf{Language datasets for vision.}
Early language datasets for vision include Flickr8K~\cite{hodosh2013framing} and Flickr30K~\cite{young2014image}.
Inspired by them, Chen \etal built a larger MS-COCO Caption~\cite{chen2015microsoft} dataset. They selected 164,062 images from MS-COCO~\cite{lin2014microsoft} and labeled each image with five sentences from independent labelers.
Recently, Visual Genome~\cite{krishna2016visual} dataset was proposed by Krishna \etal, which incorporates dense annotations of objects, attributes, and relationships within each image. 
However, although there are persons in the datasets, they are not the main subjects for descriptions and cannot be used to train person search algorithms with language descriptions.
For fine-grained visual descriptions, Reed \etal added language annotations to Caltech-UCSD birds~\cite{welinder2010caltech} and Oxford-102 flowers~\cite{nilsback2008automated} datasets to describe contents of images for text-image joint embedding.

\textbf{Deep language models for vision.}
Different from convolutional neural network which works well in image classification~\cite{krizhevsky2012imagenet,he2016deep} and object detection~\cite{kang2016object,kang2016t,kang2017object}, recurrent neural network is more suitable in processing sequential data. A large number of deep models for vision tasks~\cite{xu2015show,antol2015vqa,hu2016segmentation,johnson2015densecap,gao2015you,chen2014learning,fang2015captions} have been proposed in recent years. For image captioning, Mao \etal~\cite{mao2014deep} learned feature embedding for each word in a sentence, and connected it with the image CNN features by a multi-modal layer to generate image captions. Vinyal \etal \cite{vinyals2015show} extracted high-level image features from CNN and fed it into LSTM for estimating the output sequence. The NeuralTalk~\cite{karpathy2015deep} looked for the latent alignment between segments of sentences and image regions in a joint embedding space for sentence generation.

Visual QA methods were proposed to answer questions about given images
\cite{ren2015exploring,noh2015image,yang2015stacked,saito2016dualnet,malinowski2015ask,fukui2016multimodal}. Yang \etal~\cite{yang2015stacked} presented a stacked attention network that refined the joint features by recursively attending question-related image regions, which leads to better QA accuracy. Noh \etal \cite{noh2015image} learned a dynamic parameter layer with hashing techniques, which adaptively adjusts image features based on different questions for accurate answer classification.

Visual-semantic embedding methods~\cite{frome2013devise,karpathy2015deep,reed2016learning,liu2015multi,ren2015image} learned to embed both language and images into a common space for image classification and retrieval.
Reed \etal \cite{reed2016learning} trained an end-to-end CNN-RNN model which jointly embeds the images and fine-grained visual descriptions into the same feature space for zero-shot learning. Text-to-image retrieval can be conducted by calculating the distances in the embedding space. Frome \etal~\cite{frome2013devise} associated semantic knowledge of text with visual objects by constructing a deep visual-semantic model that re-trained the neural language model and visual object recognition model jointly.

\section{Benchmark for person search with natural language description}

Since there is no existing language dataset focusing on person appearance,
we build a large-scale benchmark for person search with natural language, termed as CUHK Person Description Dataset (CUHK-PEDES). We collected 40,206 images of 13,003 persons from five existing person re-identification datasets, CUHK03~\cite{li2014deepreid}, Market-1501~\cite{zheng2015person}, SSM~\cite{xiao2016end}, VIPER~\cite{gray2007evaluating}, and CUHK01~\cite{li2012human}, as the subjects for language descriptions. Since persons in Market-1501 and CUHK03 have many similar samples, to balance the number of persons from different domains, we randomly selected four images for each person in the two datasets.
All the image were labeled by crowd workers from Amazon Mechanical Turk (AMT), where each image was annotated with two sentence descriptions and a total of 80,412 sentences were collected.
The dataset incorporates rich details about person appearances, actions, poses and interactions with other objects. The sentence descriptions are generally long ($>23$ words in average), and has abundant vocabulary and little repetitive information. Examples of our proposed dataset are shown in Figure~\ref{fig:dataset}.

\subsection{Dataset statistics}

The dataset consists of rich and accurate annotations with open word descriptions. There were 1,993 unique workers involved in the labeling task, and all of them have greater-than 95\% approving rates. We asked the workers to describe all important characteristics in the given images using sentences with at least 15 words. 
The large number of workers means the dataset has diverse language descriptions and methods trained with it are unlikely to overfit to descriptions of just a few workers.

\begin{figure}
\begin{center}
\includegraphics[width=0.9\linewidth]{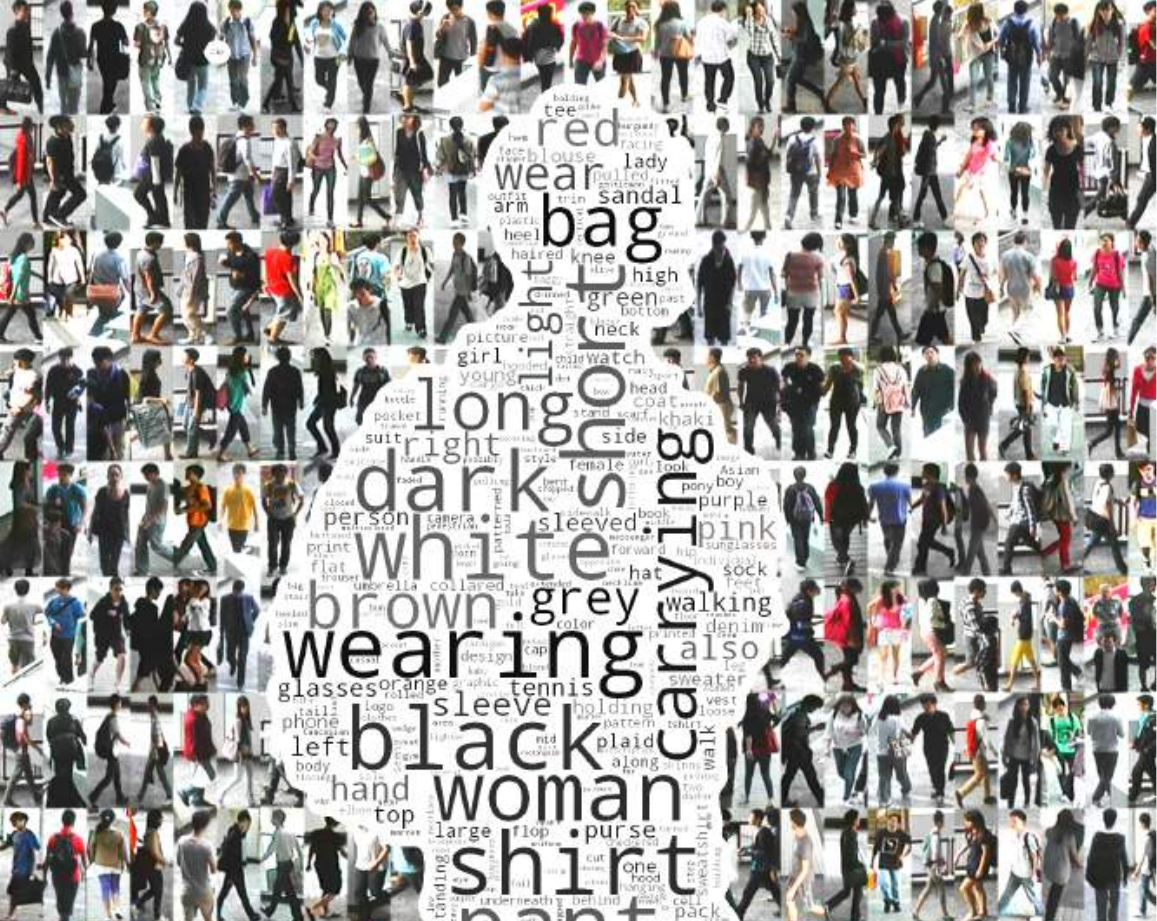} \ \\
\end{center}
\vspace{-7pt}
\caption{High-frequency words and person images in our dataset.}
\label{fig:wordcould}
\vspace{-7pt}
\end{figure}

Vocabulary, phrase sizes, and sentence length are important indicators on the capacity our language dataset. There are a total of 1,893,118 words and 9,408 unique words in our dataset. 
The longest sentence has 96 words and the average word length is 23.5 which is significantly longer than the 5.18 words of MS-COCO Caption \cite{lin2014microsoft} and the 10.45 words of Visual Genome~\cite{krishna2016visual}. Most sentences have 20 to 40 words in length. Figure~\ref{fig:wordcould} illustrates some person examples and high-frequency words.


\begin{figure}
\begin{center}
\includegraphics[width=1\linewidth]{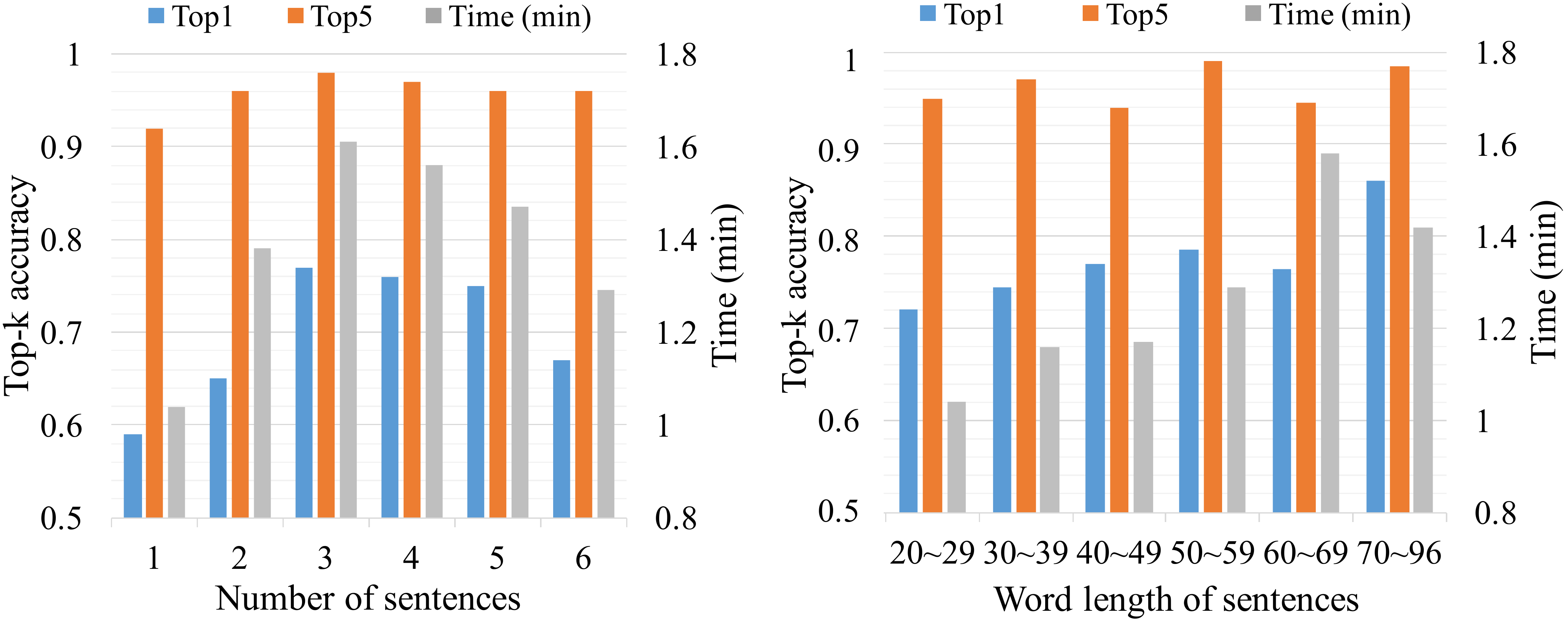} \ \\
\end{center}
\vspace{-9pt}
   \caption{Top-1 accuracy, top-5 accuracy, and average used time of manual person search using language descriptions with different number of sentences and different sentence lengths.}
\label{fig:sentnumberlength}
\end{figure}

\subsection{User study}
\label{sec:userstudy}
Based on the language annotations we collect, we conduct the user studies to investigate 1) the expressive power of language descriptions compared with that of attributes, 2) the expressive power in terms of the number of sentences and sentence length, and 3) the expressive power of different word types. The studies provide us insights for understanding the new problem and guidance on designing our neural networks.

\textbf{Language vs. attributes.} Given a descriptive sentence or annotated attributes of a query person image, we ask crowd workers from AMT to select its corresponding image from a pool of 20 images. The 20 images consist of the ground truth image, 9 images with similar appearances to the ground truth, and 10 randomly selected images from the whole dataset. The 9 similar images are chosen from the whole dataset by the LOMO+XQDA~\cite{liao2015person} method, which is a state-of-the-art method for person re-identification. The other 10 distractor images are randomly selected and have no overlap with the 9 similar images. The person attribute annotations are obtained from the PETA~\cite{deng2014pedestrian} dataset, which have 1,264 same images with our dataset. A total of 500 images are manually searched by the workers, and the average top-1 and top-5 accuracies of the searches are evaluated. The searches with language descriptions have 58.7\% top-1 and 92.0\% top-5 accuracies, while the searches with attributes have top-1 and top-5 accuracies of 33.3\% and 74.7\% respectively. In terms of the average used time for each search, using language descriptions takes 62.18$s$, while using attributes takes 81.84$s$.
The results show that, from human's perspective, language descriptions are much precise and effective in describing persons than attributes. They partially endorse our choice of using language descriptions for person search.

\textbf{Sentence number and length.} We design manual experiments to investigate the expressive power of language descriptions in terms of the number of sentences for each image and sentence length. The images in our dataset are categorized into different groups based on the number of sentences associated with each image and based on different sentence lengths. Given the sentences for each image, we ask crowd workers from AMT to manually retrieve  the corresponding images from pools of 20 images.
The average top-1 and top-5 accuracies, and used time for different image groups are shown in Figure \ref{fig:sentnumberlength}, which show that 3 sentences for describing a person achieved the highest retrieval accuracy. The longer the sentences are, the easier for users to retrieve the correct images.

\setlength{\tabcolsep}{7pt}
\begin{table}[t]
\footnotesize
\begin{center}
\begin{tabular}{c|c|c|c|c}
\hline

\hline
\cline{2-4}
& orig. sent. & w/o nouns & w/o adjs & w/o verbs \\
\hline

\hline\noalign{\smallskip}\hline

\hline
top-1 			& 0.59 	& 0.38 	& 0.44 	& 0.57 \\
\hline
top-5  			& 0.92  & 0.81 	& 0.85 	& 0.92 \\
\hline
time (min)  	& 1.14  & 1.01  & 0.98  & 1.12\\
\hline

\hline
\end{tabular}
\end{center}
\vspace{-5pt}
\caption{Top-1 accuracy, top-5 accuracy, and average used time of manual person search results using the original sentences, and sentences with nouns, or adjectives, or verbs masked out.}
\label{tab:nouns}
\vspace{-5pt}
\end{table}






\textbf{Word types.} We also investigate the importance of different word types, including nouns, verbs, and adjectives by using manual experiments with the same 20-image pools. For this study, nouns, or verbs, or adjectives in the sentences are masked out before provided to the workers. For instance, ``the girl has pink hair'' is converted to ``the **** has pink ****'', where the nouns are masked out. Results in Table~\ref{tab:nouns} demonstrate that the nouns provide most information followed by the adjectives, while the verbs carry least information. This investigation provides us important insights that nouns and adjectives should be paid much attention to when we design neural networks or collecting new language data.

\begin{figure}
\begin{center}
\includegraphics[width=0.85\linewidth]{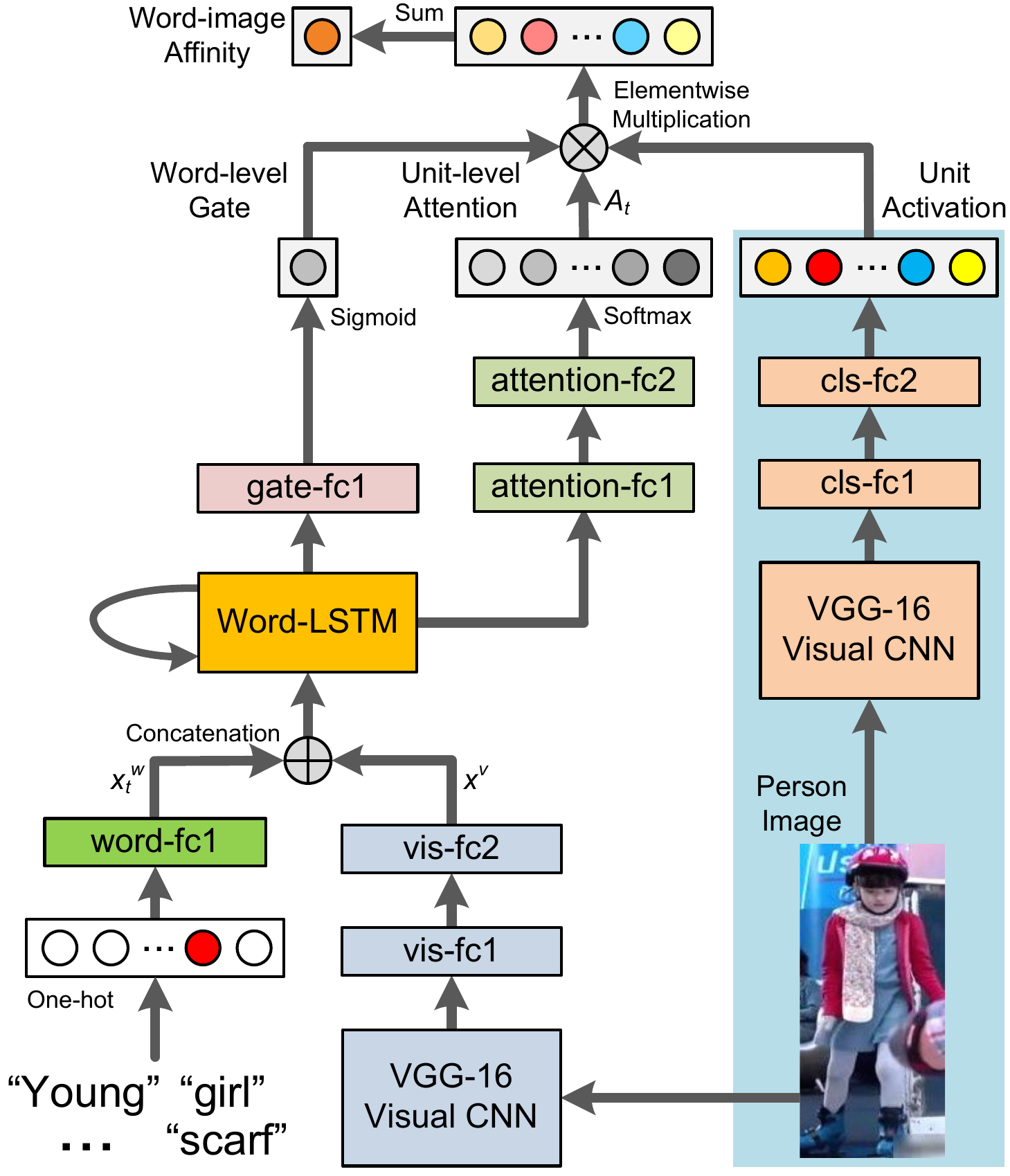} \ \\
\end{center}
\vspace{-6pt}
\caption{The network structure of the proposed GNA-RNN. It consists of a visual sub-network (right blue branch) and a language sub-network (left branch). The visual sub-network generates a series of visual units, each of which encodes if certain appearance patterns exist in the person image. Given each input word, The language sub-network outputs world-level gates and unit-level attentions for weighting visual units.}
\label{fig:GNARNN}
\vspace{-7pt}
\end{figure}

\section{GNA-RNN model for pedestrian search}

The key to address the person search with language description is to effectively build word-image relations. Given each word, it is desirable if the neural network would search related regions to determine whether the word with its context fit the image. For a sentence, all such word-image relations can be investigated, and confidences of all relations should be weighted and then aggregated to generate the final sentence-image affinity.

Based on this idea, we propose a novel deep neural network with Gated Neural Attention (GNA-RNN) to capture word-image relations and estimate the affinity between a sentence and a person image.
The overall structure of the GNA-RNN is shown in Figure~\ref{fig:GNARNN}. The network model consists of a visual sub-network and a language sub-network. The visual sub-network generates a series of visual unit activations, each of which encodes if  certain human attributes or appearance patterns (\eg, white scarf) exist in the given person image. The language sub-network is a Recurrent Neural Network (RNN) with Long Short-Term Memory (LSTM) units, which takes words and images as input. At each word, it outputs unit-level attention and word-level gate to weight the visual units from the visual sub-network. The unit-level attention determines which visual units should be paid more attention to according to the input word. 
The word-level gate weight the importance of different words. 
All units' activations are weighted by both the unit-level attentions and word-level gates, and are then aggregated to generate the final affinity. By training such network in an end-to-end manner, the Gated Neural Attention mechanism is able to effectively capture the optimal word-image relations.

\subsection{Visual units}

The visual sub-network takes person images that are resized to 256$\times$256 as inputs. It has the same bottom structure as VGG-16 network, and adds two 512-unit fully-connected layers at the ``drop7'' layer to generate 512 visual units, ${\bf v} = [v_1,...,v_{512}]^T$. Our goal is to train the whole network jointly such that each visual unit determines whether certain human appearance pattern exist in the person image. 
The visual sub-network is first pre-trained on our dataset for person classification based on person IDs. During the joint training with language sub-network, only parameters of the two new fully-connected layers (``cls-fc1'' and ``cls-fc2'' in Figure~\ref{fig:GNARNN}) are updated for more efficient training.
Note that we do not manually constrain which units learn what concepts. The semantic meanings of the visual units automatically capture necessary semantic concepts via jointly training of the whole network.

\subsection{Attention over visual units}
\label{sec:neuralattention}

To effectively capture the word-image relations, we propose a unit-level attention mechanism for visual units. At each word, the visual units having similar semantic meanings with the word should be assigned with more weights. Take Figure~\ref{fig:GNARNN} as example, given the words ``white scarf'', the language sub-network would attend more the visual unit that corresponds to the concept of ``white scarf''.
We train the language sub-network to to achieve this goal.

The language sub-network is a LSTM network~\cite{hochreiter1997long}, which is effective at capturing temporal relations of sequential data. Given an input sentence, the LSTM generates attentions for visual units word by word. The words are first encoded into length-$K$ one-hot vectors, where $K$ is the vocabulary size. Given a descriptive sentence, a learnable fully connected layer (``word-fc1'' in Figure~\ref{fig:GNARNN}) converts the $t$th raw word to a word embedding feature $x^t_w$. Two 512-unit fully connected layers (``vis-fc1'' and ``vis-fc2'' in Figure~\ref{fig:GNARNN}) following the ``drop7''  layer of VGG-16 are treated as visual features $x_v$ for the LSTM. 
At each step, the LSTM takes $x_t = [x_t^w, x^v]^T$ as input, which is concatenation of $t$th word embedding $x^w_t$ and image features $x^v$. 

The LSTM consists of a memory cell $c_t$ and three controlling gates, \ie input gate $i_t$, forget gate $f_t$, and output gate $o_t$. The memory cell preserves the knowledge of previous step and current input while the gates control the update and flow direction of information.
At each word, the LSTM updates the memory cell $c_t$ and output a hidden state $h_t$ in the following way,
\vspace{-5pt}
\begin{align}
i_t &= \sigma{(W_{xi} x_t + W_{hi} h_{t-1} + b_i)}, \nonumber \\
f_t &= \sigma{(W_{xf} x_t + W_{hf} h_{t-1} + b_f)}, \nonumber \\
o_t &= \sigma{(W_{xo} x_t + W_{ho} h_{t-1} + b_o)}, \\
c_t &= f_t \odot c_{t-1} + i_t \odot h{(W_{xc} x_t + W_{hc} h_{t-1} + b_c)}, \nonumber\\
h_t &= o_t \odot h{(c_t)}, \nonumber
\end{align}
where $\odot$ represents the element-wise multiplication, $W$ and $b$ are parameters to learn. 

For generating the unit-level attentions at each word, the output hidden state $h_t$ is fed into a fully-connected layer with ReLU non-linearity function and a fully-connected layer with softmax function to obtain the attention vector $A_t \in \mathbb{R}^{512}$, which has the same dimension as the visual units ${\bf v}$.
The affinity between the sentence and the person image at the $t$th word can then be obtained by
\vspace{-5pt}
\begin{align}
a_t =& \sum_{n=1}^{512}  A_t(n) v_n,~~~~~\textnormal{s.t.} \, \sum_{n=1}^{512} A_t(n) = 1,
\end{align}
where $A_t(n)$ denotes the attention value for the $n$th visual unit. Since each visual unit determines the existence of certain person appearance patterns in the image, the visual units alone cannot generate sentence-image affinity. The attention values $A_t$ generated by the language sub-network decides which visual units' responses should be summed up to compute the affinity value. If the language sub-network generates high attention value at certain visual unit, only if the visual unit also has high response, which denotes existence of certain visual concepts, will the elementwise multiplication generates high affinity value at this word. The final sentence-image affinity is summation of affinity values at all words, $a = \sum_{t=1}^T a_t$, where $T$ is the number of words in the given sentence.

\setlength{\tabcolsep}{7pt}
\begin{table*}[t]
\begin{center}
\begin{tabular}{c|c|cc|ccc|c}
\hline

\hline
\cline{2-8}
& NeuralTalk~\cite{vinyals2015show} & CNN-RNN~\cite{reed2016learning} & EmbBoW & QAWord & QAWord-img & QABoW & GNA-RNN\\
\hline

\hline\noalign{\smallskip}\hline

\hline
top-1 		& 13.66 & 8.07 & 8.38 & 11.62 & 10.21  & 8.00 & \textbf{19.05} \\
\hline
top-10  	& 41.72 & 32.47 & 30.76 & 42.42 & 44.53 & 30.56 & \textbf{53.64}  \\
\hline

\hline
\end{tabular}
\end{center}
\vspace{-7pt}
\caption{Quantitative results of the proposed GNA-RNN and compared methods on the proposed dataset.}
\label{tab:results}
\vspace{-5pt}
\end{table*}

\subsection{Word-level gates for visual units}
\label{sec:wordgate}

The unit-level attention is able to associate the most related units to each word. However, the attention mechanism requires different units' attentions competing with each other. In our case with the softmax non-linearity function, we have $\sum_{n=1}^{512} A_t(n) = 1$, and found that such constraints are important for learning effective attentions.

However, according to our user study on different word types in Section~\ref{sec:userstudy}, different words carry significantly different amount of information for obtaining language-image affinity. For instance, the word ``white'' should be more important than the word ``this''. At each word, the unit-level attentions always sum up to $1$ and cannot reflect such differences. Therefore, we propose to learn world-level scalar gates at each word for learning to weight different words. The word-level scalar gate is obtained by mapping the hidden state $h_t$ of the LSTM via a fully-connected layer with sigmoid non-linearity function $g_t = \sigma(W_g h_t + b_g)$,
where $\sigma$ denotes the sigmoid function, and $W_g$ and $b_g$ are the learnable parameters of the fully-connected layer.

Both the unit-level attention and world-level gate are used to weight the visual units at each word to obtain the per-word language-image affinity $\hat{a}_t$,
\begin{align} 
	\hat{a}_t = g_t \sum_{n=1}^{512}  A_t(n) v_n,
\end{align}
and the final affinity is the aggregation of affinities at all words $\hat{a} = \sum_{t=1}^T \hat{a}_t$.

\subsection{Training scheme}

The proposed GNA-RNN is trained end-to-end with batched Stochastic Gradient Descent, except for the VGG-16 part of the visual sub-network, which is pre-trained for person classification and fixed afterwards. The training samples are randomly chosen from the dataset with corresponding sentence-image pairs as positive samples and non-corresponding pairs as negative samples. The ratio between positive and negative samples is 1:3.
Given the training samples,
the training minimizes the cross-entropy loss,
\vspace{-5pt}
\begin{align}
	E = -\frac{1}{N}\sum_{i=1}^{N} \left[ y^i \log \hat{a}^i  + (1-y^i) \log (1-\hat{a}^i \right)]
\end{align}
where $\hat{a}^i$ denotes the predicted affinity for the $i$th sample, and $y^i$ denotes its ground truth label, with $1$ representing corresponding sentence-image pairs and $0$ representing non-corresponding ones.
We use 128 sentence-image pairs for each training batch. All fully connected layers except for the one for word-level gates have 512 units.

\section{Experiments}

There is no existing method specifically designed for the problem. We investigate a wide range of possible solutions based on state-of-the-art language models for vision tasks, and compare those solutions with our proposed method. We also conduct component analysis of our proposed deep neural networks to show that our proposed Gated Neural Attention mechanism is able to capture complex word-image relations. Extensive experiments and comparisons with state-of-the-art methods demonstrate the effectiveness of our GNA-RNN for this problem.

\subsection{Dataset and evaluation metrics}
\label{exp:dataset}
The dataset is splitted into three subsets for training, validation, and test without having overlaps with same person IDs. The training set consists of 11,003 persons, 34,054 images and 68,108 sentence descriptions. The validation set and test set contain 3,078 and 3,074 images, respectively, and both of them have 1,000 persons. All experiments are performed based on this train-test split.

We adopt the top-$k$ accuracy to evaluate the performance of person retrieval. Given a query sentence, all test images are ranked according to their affinities with the query. A successful search is achieved if any image of the corresponding person is among the top-$k$ images. Top-$1$ and top-$10$ accuracies are reported for all our experiments.

\setlength{\tabcolsep}{2pt}
\begin{table}[t]
\begin{center}
\begin{tabular}{c|c|c|c|c}
\hline

\hline
\cline{2-4}
& GNA-RNN & w/o pre-train & w/o gates & w/o attention \\
\hline

\hline\noalign{\smallskip}\hline

\hline
top-1 		& \textbf{19.05} &8.93 & 13.86 & 4.85 \\
\hline
top-10  	& \textbf{53.64} &32.32 & 44.27 & 27.16 \\
\hline

\hline
\end{tabular}
\end{center}
\vspace{-5pt}
\caption{Quantitative results of GNA-RNN on the proposed dataset without VGG-16 re-id pre-training, without world-level gates or without unit-level attentions.}
\label{tab:ablation}
\vspace{-5pt}
\end{table}

\setlength{\tabcolsep}{7pt}
\begin{table}[t]
\begin{center}

\begin{tabular}{c|c|c|c|c|c}
\hline

\hline
\# units & 128 & 256 & 512 & 1024 & 2048\\
\hline

\hline\noalign{\smallskip}\hline

\hline
top-1 		& 16.15 & 16.75 & 19.05 & 18.62 & 18.25 \\
\hline
top-10  	& 48.58 & 49.25 & 53.64 & 52.39 & 51.59 \\
\hline

\end{tabular}
\end{center}
\vspace{-5pt}
\caption{Top-1 and top-10 accuracies of GNA-RNN with different number of visual units.}
\label{tab:VN}
\vspace{-10pt}
\end{table}


\subsection{Compared methods and baselines}
\label{exp:baseline}

We compare a wide range of possible solutions with deep neural networks, including methods for image captioning, visual QA, and visual-semantic embedding. 
Generally, each type of methods utilize different supervisions for training. Image captioning, visual QA, and visual-semantic embedding methods are trained with word classification losses, answer classification losses, and distance-based losses, respectively.
We also propose several baselines to investigate the influences of detailed network structure design. To make fair comparisons, the image features for all compared methods are from our VGG-16 network pre-trained model.

\textbf{Image captioning.}
Vinyals \etal~\cite{vinyals2015show} and Karpathy \etal~\cite{karpathy2015deep} proposed to generate natural sentences describing an image using deep recurrent frameworks. We use the code provided by Karpathy \etal to train the image captioning model. We follow the testing strategy in~\cite{hu2015natural} to use image captioning method for text-to-image retrieval.
During the test phase, given a person image, instead of recursively using the predicted word as inputs of the next time step to predict the image caption, the LSTM takes the given sentence word by word as inputs. It calculates the per-word cross entropy losses between the given word and the predicted word from LSTM. Corresponding sentence-image pairs would have low average losses, while non-corresponding ones would have higher average losses.

\textbf{Visual QA.}
Agrawal \etal \cite{antol2015vqa} proposed the deeper LSTM Q + norm I method to answer questions about the given image. We replace the element-wise multiplication between the question and image features, with concatenation of question and image features, and replace the multi-class classifier with a binary classifier. 
Since the proposed GNA-RNN has only one layer for the LSTM, we change the LSTM in deeper LSTM Q + norm I to one layer as well for fair comparison. The norm I in \cite{antol2015vqa} is also changed to contain two additional fully-connected layers to obtain image features instead of the original one layer following our model's structure.
We call the modified model QAWord.
Where to concatenate features of question and image modalities might also influence the classification performance. The QAWord model concatenates image features with sentence features output by the LSTM. We investigate concatenating the word embedding features and image features before inputting them into the LSTM. Such a modified network is called QAWord-img. We also replace the language model in QAWord with the simple language model in \cite{zhou2015simple}, which encodes sentences using the traditional Bag-of-Word (BoW) method, and call it QABoW. 

\textbf{Visual-semantic embedding.}
These methods try to map image and sentence features into a joint embedding space.
Distances between image and sentence features in the joint space could then be interpreted as the affinities between them.
Distances between corresponding sentence-image pairs should be small, and should be high between non-corresponding paris.
Reed \etal~\cite{reed2016learning} presented a CNN-RNN for zero-shot text-to-image retrieval. We utilize their code and compare it with our proposed framework. We also investigate replacing the language model in CNN-RNN with the simple BoW language model \cite{zhou2015simple} for sentence encoding and denote it as EmbBoW.

\begin{figure*}[t]
\begin{center}
\includegraphics[width=0.96\linewidth]{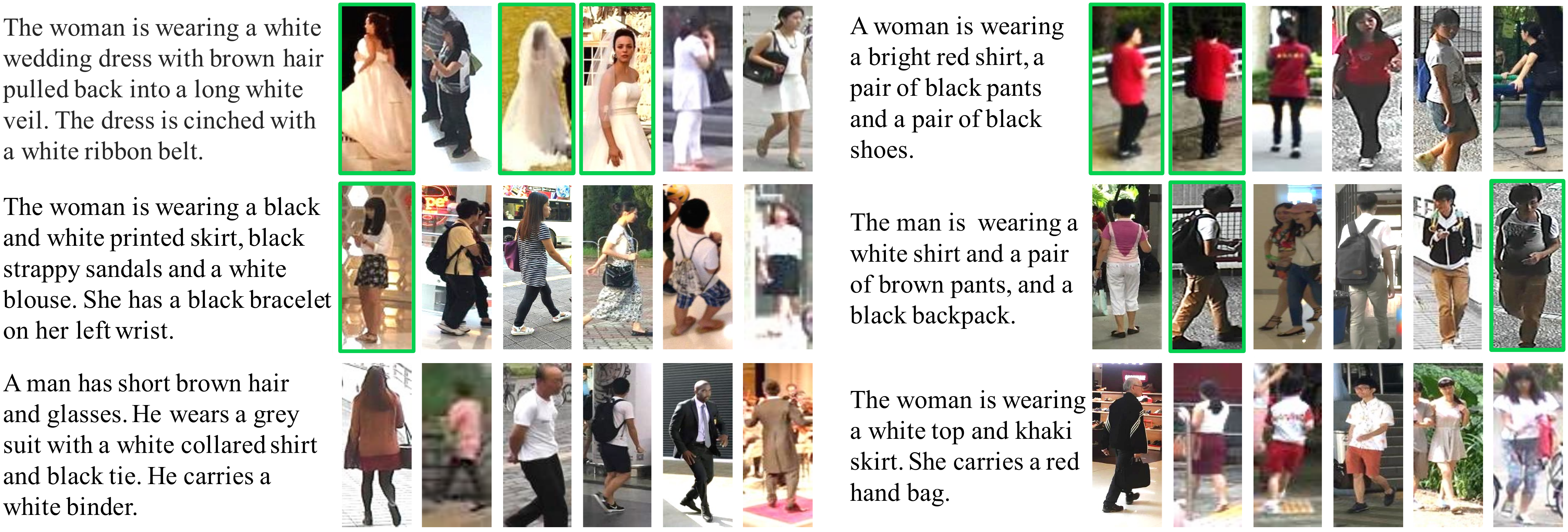}
\vspace{-7pt}
\end{center}
\caption{Examples of top-6 person search results with natural language description by our proposed GNA-RNN. Corresponding images are marked by green rectangles. (Rows 1-2) Successful searches where corresponding persons are in the top-6 results. (Row 3) Failure cases where corresponding persons are not in the top-6 results.}
\label{fig:finalexamples}
\vspace{-7pt}
\end{figure*}

\subsection{Quantitative and qualitative results}
\label{exp:results}

{\bf Quantitative evaluation.} Table~\ref{tab:results} shows the results of our proposed framework and the compared methods. We use a single sentence as query to do the person search.
Our approach achieves the best performance in terms of both top-$1$ and top-$10$ accuracies and outperforms other methods by a large margin. It demonstrates that our proposed network can better capture complex word-image relations than the compared ones.

For all the baselines, the image captioning method NeuralTalk outperforms the other baselines. It calculates the average loss at each word as the sentence-image affinity, and obtains better results than visual QA and visual embedding approaches, which encode the entire sentence into a feature vector.
Such results show that the LSTM might have difficulty encoding complex person descriptive sentences into a single feature vector. Word-by-word processing and comparison might be more suitable for the person search problem.
We also observe that QAWord-img and QAWord has similar performance. This demonstrates that, the modality fusion between image and word before or after LSTM has little impact on the person search performance. Both ways capture word-image relations to some extent. 
For the visual-semantic embedding method, the CNN-RNN does not perform well in terms of top-$k$ accuracies with the provided code. The distance-based losses might not be suitable for learning good models for the person search problem. EmbBoW and QABoW use the traditional Bag-of-Word method to encode sentences and have worse performances than their counterparts with RNN language models, which show that the RNN framework is more suitable in processing natural language data.

{\bf Component analysis.} 
We pre-train the visual VGG model for person re-id task first, and then fine-tune whole network for text-to-person search. Without the person re-id pre-training, top-1 and top-10 accuracies drop apparently as shown in Table \ref{tab:ablation}. This means the initial training affects the final performance a lot.
To investigate the effectiveness of the proposed unit-level attentions and word-level gates, we design two baselines for comparison. For the first baseline (denoted as ``w/o gates''), we remove the word-level gates and only keep the unit-level attentions. In this case, different words are equally weighted in estimating the sentence-image affinity. For the second baseline (denoted as ``w/o attention''), we try to keep the world-level gates, and replace the unit-level attentions with average pooling over units.
We list top-$1$ and top-$10$ accuracies of the two baselines in Table \ref{tab:ablation}.
Both the unit-level attention and word-level gates are important for achieving good performance by our GNA-RNN.

{\bf Investigation on the impact of the number of visual units.} 
Results of different number of visual units are listed in Table \ref{tab:VN}. Models with more visual units might over-fit the dataset. 512 units achieves the best result.

{\bf Qualitative evaluation.} We conduct qualitative evaluation for our proposed GNA-RNN. Figure \ref{fig:finalexamples} shows 6 person search results with natural language descriptions by our proposed GNA-RNN. The four cases in the top 2 rows show successful cases where corresponding images are within the top-6 retrieval results. For the successful cases, we can observe that each top image has multiple regions that fit parts of the descriptions. Some non-corresponding images also show correlations to the query sentences. In terms of failure cases, there are two types of them. The first type of failure searches do retrieve images that are similar to the language descriptions, however, the exact corresponding images are not within the top retrieval results. For instance, the bottom right case in Figure \ref{fig:finalexamples} does include persons (top-2, top-3, and top-4) similar to the descriptions, who all wear white tops and red shorts/skirts. Other persons have some characteristics that partially fits the descriptions. The top-1 person has a ``hand bag''. The top-4 person wears ``white top'', and the top-6 person carries  a ``red bag''. The second type of failure cases show that the GNA-RNN fails to understand the whole sentence but only captures separate words or phrases. Take the bottom left case in Figure \ref{fig:finalexamples} as an example, the phrase ``brown hair'' is not encoded correctly. Instead, only the word ``brown'' is captured, which leads to the ``brown'' suit for the top-1 and top-6 persons, and ``brown'' land in the top-2 image. We also found some rare words/concepts or detailed descriptions are difficult to learn and to locate, such as ``ring'', ``bracelet'', ``cell phones'', \etc, which might be learned if more data is provided in the future.

\begin{figure}
\begin{center}
\includegraphics[width=0.98\linewidth]{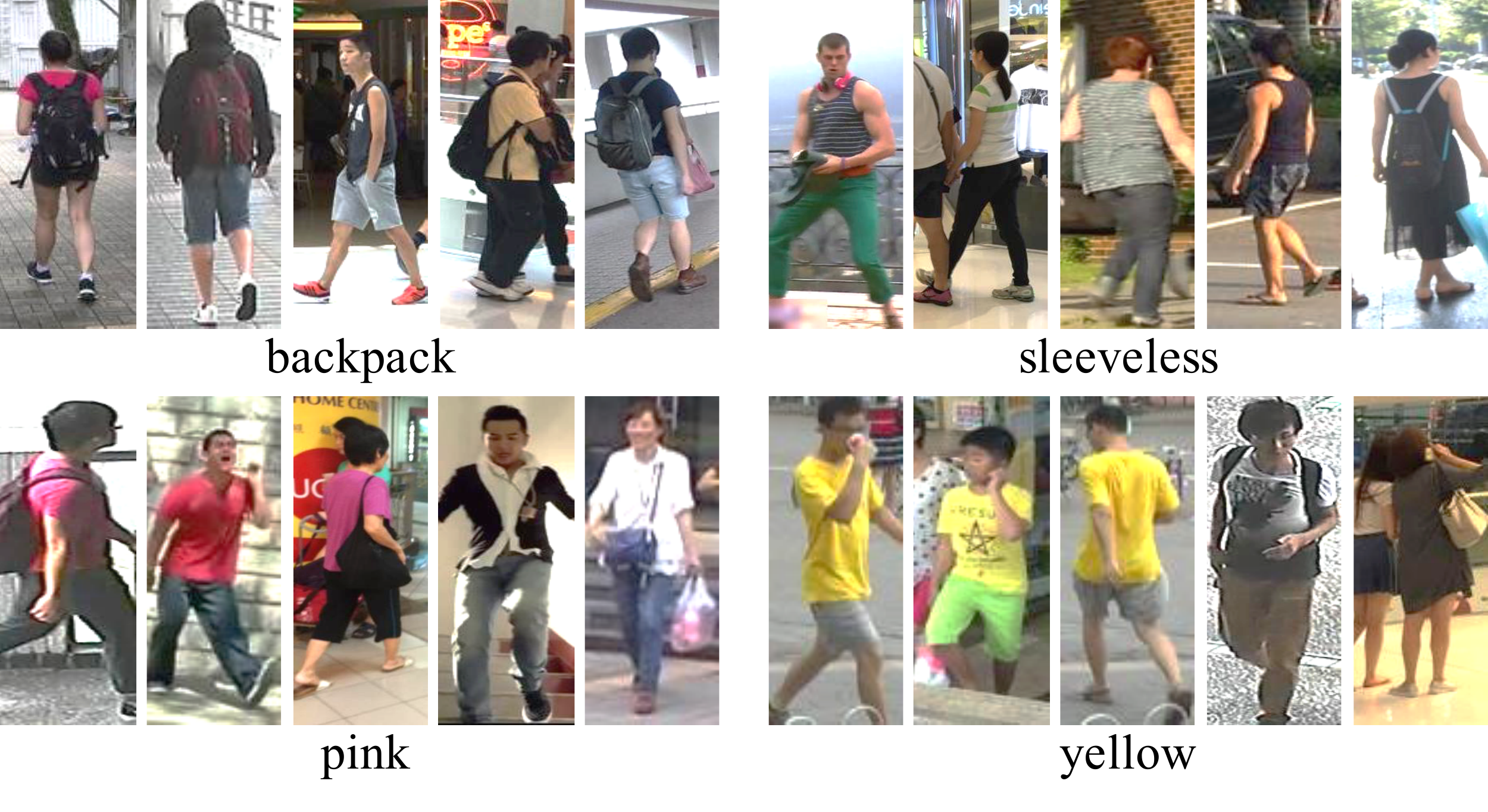}
\end{center}
\vspace{-9pt}
\caption{Images with the highest activations on 4 different visual units. The 4 units are identified as the one with the maximum average attention values in our GNA-RNN with the same word (``backpack'', ``sleeveless'', ``pink'', ``yellow'') and a large number of images. Each unit determines the existence of some common visual patterns.}
\label{fig:units}
\vspace{-8pt}
\end{figure}

{\bf Visual unit visualization.} We also inspect the learned visual units to see whether they implicitly capture common visual patterns in person images. We choose some frequent adjectives and nouns. For each frequent word, we collect its unit-level attention vectors for a large number of training images. Such unit-level attention vectors are averaged to identify its most attended visual units. For each of such units, we retrieve the training images that have the highest responses on the units. Some examples of the visual units obtained in this way are shown in Figure \ref{fig:units}. Each of them captures some common image patterns.
 
\section{Conclusions}

In this paper, we studied the problem of person search with natural languages. We collected a large-scale person dataset with 80,412 sentence descriptions of 13,003 persons. Various baselines are evaluated and compared on the benchmark. A GNA-RNN model was proposed to learn affinities between sentences and person images with the proposed gated neural attention mechanism, which established the state-of-the art performance on person search.


\vspace{5pt}
\noindent
{\bf Acknowledgement} 
This work is supported in part by SenseTime Group Limited, in part by the General Research Fund through the Research Grants Council of Hong Kong under Grants CUHK14207814, CUHK14213616, CUHK14206114, CUHK14205615, CUHK419 412, CUHK14203015 and CUHK14239816, in part by the Hong Kong Innovation and Technology Support Programme Grant ITS/121/15FX, in part by National Natural Science Foundation of China  under Grant 61371192, in part by the Ph.D. Program Foundation of China under Grant 20130185120039, and in part by the China Postdoctoral Science Foundation under Grant 2014M552339.

{\small
\bibliographystyle{ieee}
\bibliography{egbib}
}

\end{document}